%% file: main.tex
\documentclass[10pt,twocolumn,letterpaper]{article}

\usepackage[pagenumbers]{iccv} 


\definecolor{iccvblue}{rgb}{0.21,0.49,0.74}
\usepackage[pagebackref,breaklinks,colorlinks,allcolors=iccvblue]
{hyperref}
\usepackage{array}
\usepackage{makecell}
\usepackage{multirow}
\usepackage{booktabs}
\usepackage{pifont}
\usepackage{placeins}
\usepackage{gensymb}
\usepackage{graphicx}
\usepackage{subcaption}
\expandafter\def\csname ver@subfig.sty\endcsname{}
\usepackage{listings}
\usepackage[most]{tcolorbox} 
\definecolor{spatialblue}{rgb}{0.05,0.35,0.85}
\usepackage{tikz}
\definecolor{softgray}{rgb}{0.5,0.5,0.5}

\def\paperID{5952} 
\def\confName{ICCV}
\def\confYear{2025}

\title{\textsc{NuScenes-SpatialQA}: A Spatial Understanding and Reasoning Benchmark for Vision-Language Models in Autonomous Driving}

\author{
\makebox[\textwidth][c]{%
\begin{tabular}{c}
\textbf{
Kexin Tian$^{1}$ \hspace{0.5em}
Jingrui Mao$^{1}$ \hspace{0.5em}
Yunlong Zhang$^{1}$ \hspace{0.5em}
Jiwan Jiang$^{2}$ \hspace{0.5em}
Yang Zhou$^{1}$\thanks{Corresponding Authors.} \hspace{0.5em}
Zhengzhong Tu$^{1}$\footnotemark[1]
} \\[0.2em]
$^{1}$Texas A\&M University \quad $^{2}$University of Wisconsin-Madison \\
{\tt\small \{ktian6, yangzhou295, tzz\}@tamu.edu}\\[1.0em]
\href{https://taco-group.github.io/NuScenes-SpatialQA/}{\textcolor{iccvblue}{taco-group.github.io/NuScenes-SpatialQA/}}
\end{tabular}
}
}

\begin{document}
\maketitle
\input{sec/0_abstract}    
\input{sec/1_introduction}

\input{sec/2_related_works}

\input{sec/3_methodology}
\input{sec/4_experiments}

\input{sec/5_conclusions}

{
    \small
    \bibliographystyle{ieeenat_fullname}
    \bibliography{main}
}
\input{sec/6_supplementary}

\end{document}

%% file: sec/0_abstract.tex
\begin{abstract}

Recent advancements in Vision-Language Models (VLMs) have demonstrated strong potential for autonomous driving tasks. However, their spatial understanding and reasoning—key capabilities for autonomous driving—still exhibit significant limitations. Notably, none of the existing benchmarks systematically evaluate VLMs' spatial reasoning capabilities in driving scenarios. To fill this gap, we propose \textbf{NuScenes-SpatialQA}, the first large-scale ground-truth-based Question-Answer (QA) benchmark specifically designed to evaluate the spatial understanding and reasoning capabilities of VLMs in autonomous driving. Built upon the NuScenes dataset, the benchmark is constructed through an automated 3D scene graph generation pipeline and a QA generation pipeline. The benchmark systematically evaluates VLMs' performance in both spatial understanding and reasoning across multiple dimensions. Using this benchmark, we conduct extensive experiments on diverse VLMs, including both general and spatial-enhanced models, providing the first comprehensive evaluation of their spatial capabilities in autonomous driving. Surprisingly, the experimental results show that the spatial-enhanced VLM outperforms in qualitative QA but does not demonstrate competitiveness in quantitative QA. In general, VLMs still face considerable challenges in spatial understanding and reasoning. 

\end{abstract}

%% file: sec/1_introduction.tex
\section{Introduction}

\begin{figure}[h]
    \centering
    \includegraphics[width=0.9\linewidth]{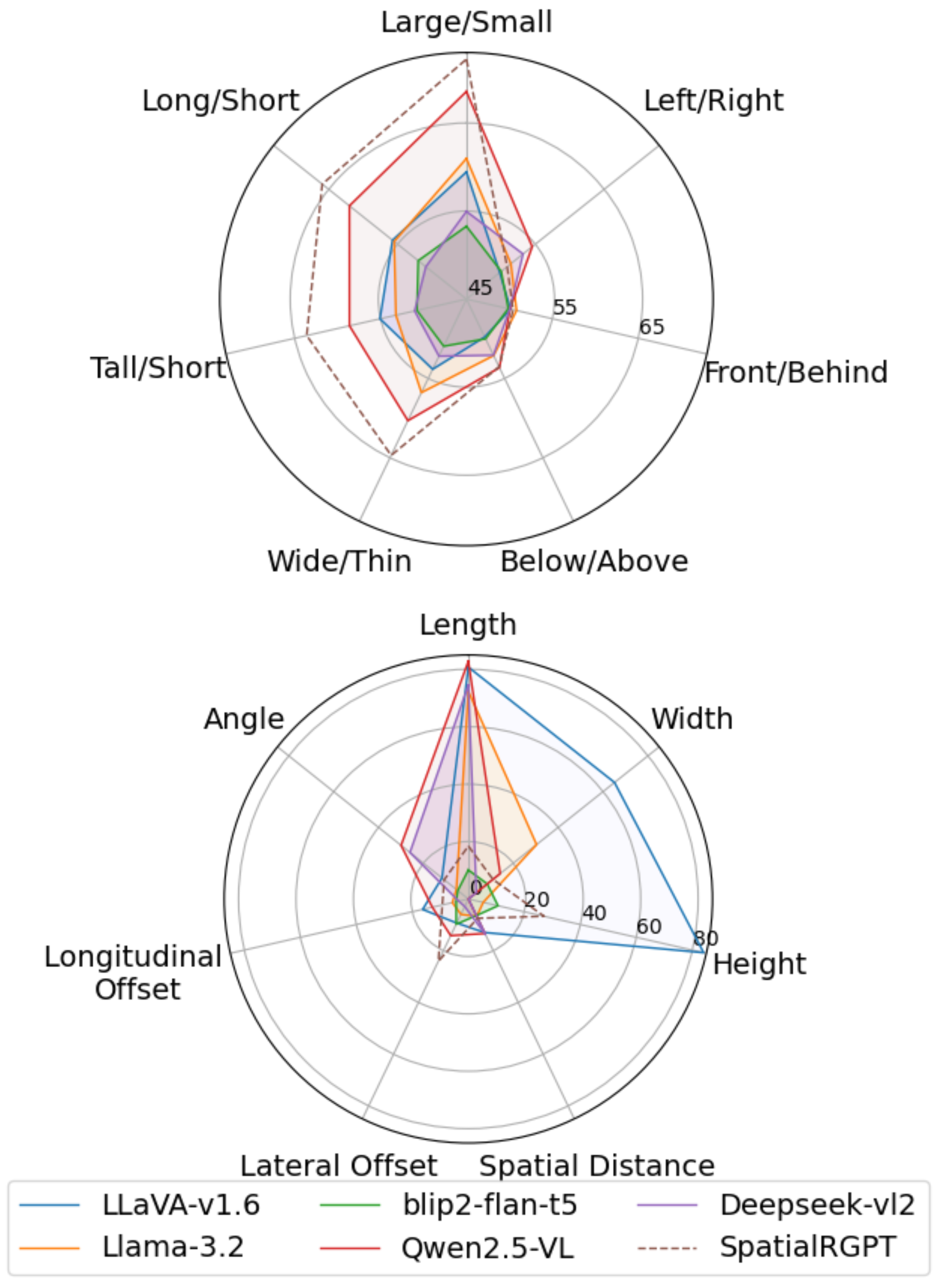}
    \label{fig:quantitative}
    \caption{
    Comprehensive experiments on our NuScenes-SpatialQA benchmark have demonstrated VLMs' performance on spatial understanding and reasoning abilities, including spatial relationship tasks (top) and quantitative spatial measurement tasks (bottom).\S}
    \label{fig:qualitative_quantitative}
\end{figure}

Vision-Language Models (VLMs)~\cite{liu2023visual, meta2024llama, bai2025qwen2, li2023blip2bootstrappinglanguageimagepretraining, wu2024deepseekvl2mixtureofexpertsvisionlanguagemodels} have made remarkable progress in recent years, demonstrating strong performance across diverse vision-language tasks, including image captioning~\cite{guo2024regiongpt, alaluf2024myvlm}, visual question answering~\cite{zheng2024simplellm4adendtoendvisionlanguagemodel, ghosal2023languageguidedvisualquestion}, and visual grounding~\cite{Kuckreja_2024_CVPR}.
Leveraging these capabilities, VLMs are increasingly recognized for their potential to significantly enhance scene understanding and reasoning, which is especially notable in multimodal perception and reasoning contexts, such as autonomous driving~\cite{jiang2024sennabridginglargevisionlanguage,zheng2024simplellm4adendtoendvisionlanguagemodel, tian2024drivevlm, gopalkrishnan2024multi, nie2024reason2drive, xing2025openemma, yuan2024rag, wang2024omnidrive, gao2025stamp}. Consequently, recent works have applied VLMs to various driving-relevant tasks such as object recognition~\cite{hwang2024emma,xing2025openemmaopensourcemultimodalmodel}, scene description~\cite{wang2024omnidrive} and reasoning over driving environments~\cite{nie2024reason2drive, yuan2024rag}. 

Despite these advances, current VLMs continue to exhibit significant limitations in spatial understanding and reasoning---a critical capability for autonomous driving. Prior studies~\cite{fu2024blink} illustrate that even basic spatial tasks, such as relative depth estimation, remain substantial challenges for VLMs, underscoring a fundamental gap.
The spatial capability of VLMs influences their ability to accurately understand object relationships and determine the relative positions and distances of surrounding agents~\cite{10727090,unger2023multicamerabirdseyeview}. These abilities, in turn, impact the performance of core perception tasks, which subsequently affect downstream decision-making, including navigation, obstacle avoidance, and interaction with dynamic traffic agents~\cite{chowdhury2024deepattentiondrivenreinforcement, wang2024omnidrive, li2024sequencing}. Therefore, evaluating VLMs' spatial capability is crucial.

While several Visual Question Answering (VQA) benchmarks~\cite{cai2024spatialbotprecisespatialunderstanding, cheng2024spatialrgptgroundedspatialreasoning, chen2024spatialvlmendowingvisionlanguagemodels,7298655, hong20233dconceptlearningreasoning, hong20233dllminjecting3dworld} exist to evaluate the spatial understanding and reasoning abilities of VLMs, their applicability to autonomous driving remains limited. 
Prevailing benchmarks either focus on simplified indoor or daily-life scenes~\cite{7298655, hong20233dconceptlearningreasoning, chen2024spatialvlmendowingvisionlanguagemodels, hong20233dllminjecting3dworld}, where spatial relationships are relatively simple. 
Other benchmarks~\cite{cheng2024spatialrgptgroundedspatialreasoning, chen2024spatialvlmendowingvisionlanguagemodels, cai2024spatialbotprecisespatialunderstanding} cover a broader range of images, including outdoor and road scenes, but contain limited driving-specific scenarios, making them insufficient for systematically assessing spatial understanding in autonomous driving. Moreover, most existing spatial benchmarks~\cite{cheng2024spatialrgptgroundedspatialreasoning, chen2024spatialvlmendowingvisionlanguagemodels, hong20233dconceptlearningreasoning, hong20233dllminjecting3dworld} rely on depth estimation models such as Metric3Dv2~\cite{Hu_2024} or simulation~\cite{sun2022directvoxelgridoptimization} to approximately annotate spatial relationships. 
These external modules can introduce biases and inaccuracies that will compromise evaluation reliability. Notably, depth estimation models are known to be particularly unreliable for long-distance depth perception in outdoor driving scenes~\cite{cheng2023understandingdepthmapprogressively}, further limiting their applicability for precise spatial reasoning.While a couple of autonomous driving-focused benchmarks~\cite{qian2024nuscenesqamultimodalvisualquestion, inoue2023nuscenesmqaintegratedevaluationcaptions, sima2024drivelm, marcu2312lingoqa, xing2024autotrust} have also emerged, they do not explicitly target spatial reasoning capabilities, highlighting a clear gap in existing evaluation resources.

To bridge this gap, we introduce \textbf{NuScenes-SpatialQA}, the first-of-its-kind benchmark explicitly designed to systematically evaluate the spatial understanding and reasoning capabilities of VLMs in autonomous driving. 
Built upon the NuScenes dataset~\cite{nuscenes}, which offers extensive real-world driving scenarios with multi-modal sensor data, our NuScenes-SpatialQA primarily consists of two core components: \ding{182} a \underline{3D scene graph generation pipeline}, which automatically constructs a 3D scene graph for each scene by encoding all necessary spatial relationships between objects, and \ding{183} the \underline{QA Generation Pipeline}, which formulates question-answer pairs based on the structured 3D scene graphs. The benchmark ultimately consists of two levels of spatial questions: Spatial Understanding, which assesses the ability to directly recognize spatial properties, and Spatial Reasoning, which requires multi-hop inference beyond explicit information. To ensure highly accurate spatial representations, our benchmark utilizes ground-truth spatial information obtained from LiDAR. This provides an unbiased evaluation framework, ensuring reliable assessment of VLM performance. 

To systematically evaluate the spatial reasoning capabilities of VLMs, we conduct experiments on NuScenes-SpatialQA with both general VLMs and spatially enhanced VLM. While VLMs demonstrate moderate performance in qualitative spatial understanding, they exhibit significant limitations in quantitative tasks, with substantial variance across models. Notably, spatially enhanced VLMs surpass general VLMs in qualitative tasks but show no clear advantage in quantitative evaluation. For spatial reasoning, VLMs perform better in situational reasoning, which relies on contextual cues, than in direct spatial reasoning, which requires precise geometric inference.
In general, our contributions can be summarized as follows:
\begin{itemize}
    \item We propose \textbf{NuScenes-SpatialQA}, the first benchmark designed to evaluate VLMs' performance in both spatial understanding and spatial reasoning in autonomous driving. Our benchmark is built upon ground-truth real-world spatial data, enabling precise evaluation. 
    \item We introduce automated pipelines that generates 3D scene graphs and QA pairs from any keyframe in the nuScenes dataset. Additionally, our evaluation process does not rely on external LLM-based scorers, improving reproducibility and reducing evaluation costs.
    \item We conduct systematic experiments on multiple VLMs, analyzing their spatial reasoning capabilities in autonomous driving scenarios. Our results provide key insights into the strengths and limitations of VLMs, establishing a solid foundation for future research.
\end{itemize}

%% file: sec/2_related_works.tex
\section{Related Works}
\begin{figure*}[tbh]
    \centering
    \includegraphics[width=0.92\textwidth]{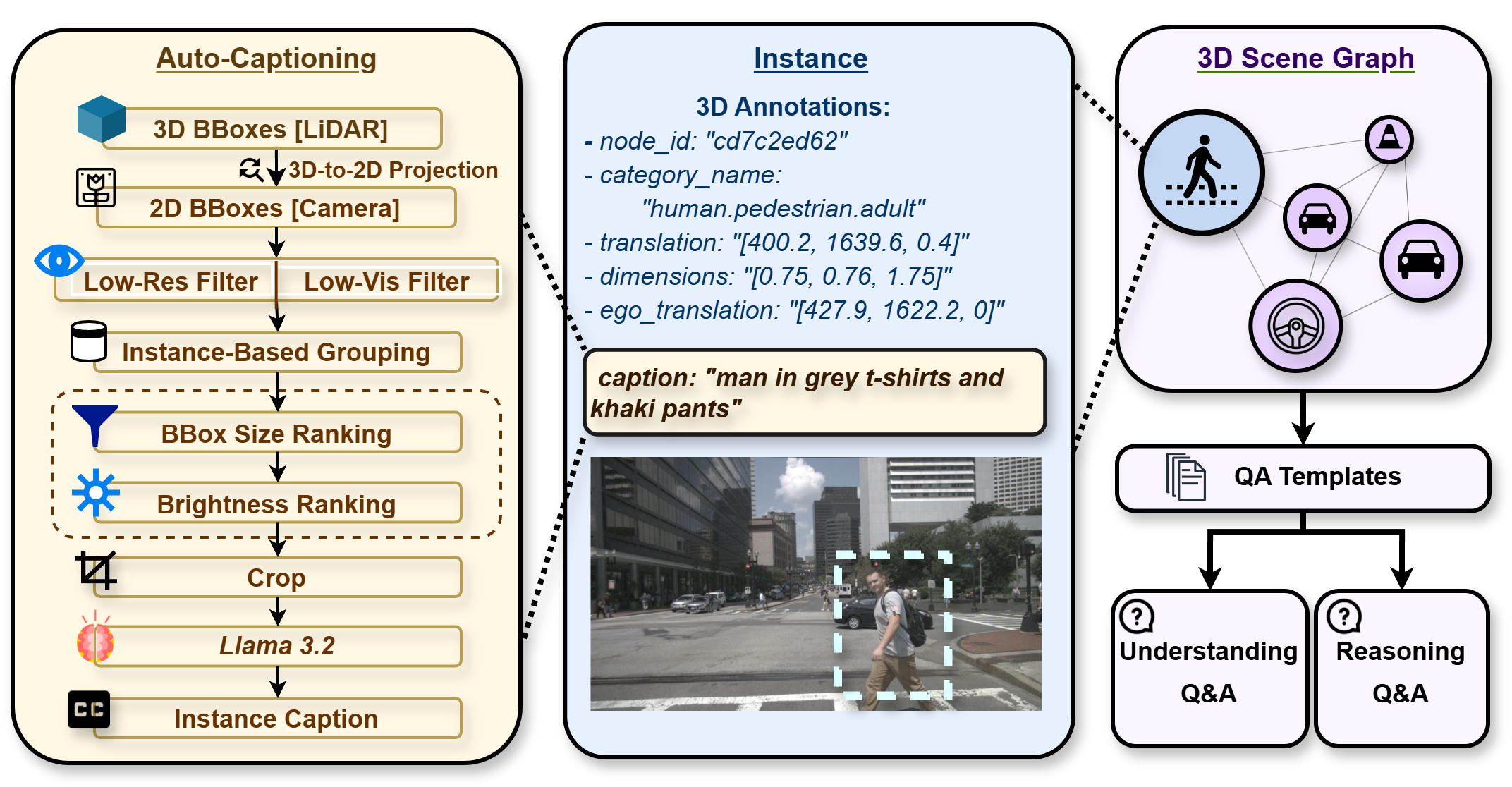}
    \vspace{-3mm}
    \caption{Overall framework of NuScenes-SpatialQA. The framework consists of two automated pipelines: (1) \textbf{Scene Graph Generation}, where a 3D scene graph is constructed using 3D annotations and instance-level captions generated from the auto-captioning process, and (2) \textbf{QA Generation}, where the constructed 3D scene graph is utilized to generate spatial question-answer pairs based on QA templates.}

    \label{fig:framework}
\end{figure*}
\paragraph{Vision-Language Models for Spatial Understanding}

In recent years, VLMs have achieved substantial advancements, leveraging large-scale multimodal pretraining to enhance their ability to interpret and generate text grounded in visual inputs. These models, including GPT-4o~\cite{openai2024gpt4ocard}, LLaVA~\cite{liu2023visual}, BLIP-2~\cite{li2023blip2bootstrappinglanguageimagepretraining}, Qwen~\cite{bai2025qwen2}, DeepSeek~\cite{wu2024deepseekvl2mixtureofexpertsvisionlanguagemodels}, CLIP~\cite{radford2021learningtransferablevisualmodels}, and Flamingo~\cite{alayrac2022flamingovisuallanguagemodel}, have demonstrated strong generalization capabilities by learning associations between textual descriptions and visual concepts. However, despite their broad applicability, these models struggle with spatial understanding and reasoning. To address these limitations, recent studies have introduced VLMs explicitly designed for spatial reasoning. SpatialVLM~\cite{chen2024spatialvlmendowingvisionlanguagemodels} and SpatialRGPT~\cite{cheng2024spatialrgptgroundedspatialreasoning} incorporate additional modules for spatial information processing and are fine-tuned on spatial datasets to enhance their reasoning capabilities. In this work, we evaluate several commonly used open-source VLMs on spatial reasoning tasks to assess their capabilities in real-world driving scenarios.

\paragraph{Benchmarks for Spatial Question Answering}

VQA is a fundamental task in vision-language research, requiring models to answer questions about images by integrating visual and textual information. To assess VLMs across diverse reasoning challenges, VQA benchmarks such as VQA-v2~\cite{goyal2017makingvvqamatter}, GQA~\cite{hudson2019gqanewdatasetrealworld}, and OK-VQA~\cite{marino2019okvqavisualquestionanswering} have been widely used, covering tasks from object recognition to compositional and commonsense reasoning. While effective for general reasoning, these benchmarks offer limited assessment of spatial relationships, which are crucial for understanding complex visual scenes. To address this, spatial benchmarks incorporate structured spatial relationships into question-answering tasks. Some benchmarks, such as CLEVR~\cite{johnson2016clevrdiagnosticdatasetcompositional}, GQA-Spatial~\cite{hudson2019gqanewdatasetrealworld}, and 3D-CLR~\cite{hong20233dconceptlearningreasoning}, focus on relatively simple indoor scenarios with well-defined object layouts. Others, including spatialVQA~\cite{chen2024spatialvlmendowingvisionlanguagemodels} and SpatialRGPT-Bench~\cite{cheng2024spatialrgptgroundedspatialreasoning}, extend spatial reasoning to more complex outdoor environments with dynamic interactions and unstructured object arrangements. However, none of these benchmarks are specifically designed to assess spatial reasoning in autonomous driving scenarios.

\paragraph{Autonomous Driving Benchmarks}

Given the complexity of driving environments and the critical importance of safety~\cite{tian2024physically, li2024beyond, li4940014nonlinear}, there is a growing need to benchmark how well VLMs understand and interpret multimodal driving scenes. Several general autonomous driving VQA benchmarks~\cite{qian2024nuscenesqamultimodalvisualquestion, inoue2023nuscenesmqaintegratedevaluationcaptions, sima2024drivelm, marcu2312lingoqa, xing2024autotrust} have been introduced to evaluate VLMs in autonomous driving. NuScenes-MQA~\cite{inoue2023nuscenesmqaintegratedevaluationcaptions}, NuScenes-QA~\cite{qian2024nuscenesqamultimodalvisualquestion}, and LingoQA~\cite{marcu2312lingoqa} primarily focus on general VQA and language understanding, assessing how well models comprehend driving scenes and generate accurate responses. DriveLM~\cite{sima2024drivelm} evaluates multi-step decision-making and causal reasoning, assessing how different factors influence driving scenarios, while AutoTrust~\cite{xing2024autotrust} examines trustworthiness aspects such as safety, privacy, and robustness. Despite these benchmarks providing valuable insights into model performance, none of these benchmarks explicitly evaluate spatial reasoning in autonomous driving, leaving a gap in assessing models' ability to understand and infer spatial relationships critical for driving decisions.

%% file: sec/3_methodology.tex
\section{Methodology}

 In this section, we outline the methodology for constructing nuScenes-SpatialQA. The overall framework is shown in Figure \ref{fig:framework}. 

\subsection{Developments \label{developments}}

\begin{figure*}
    \centering
    \includegraphics[width=0.99\linewidth]{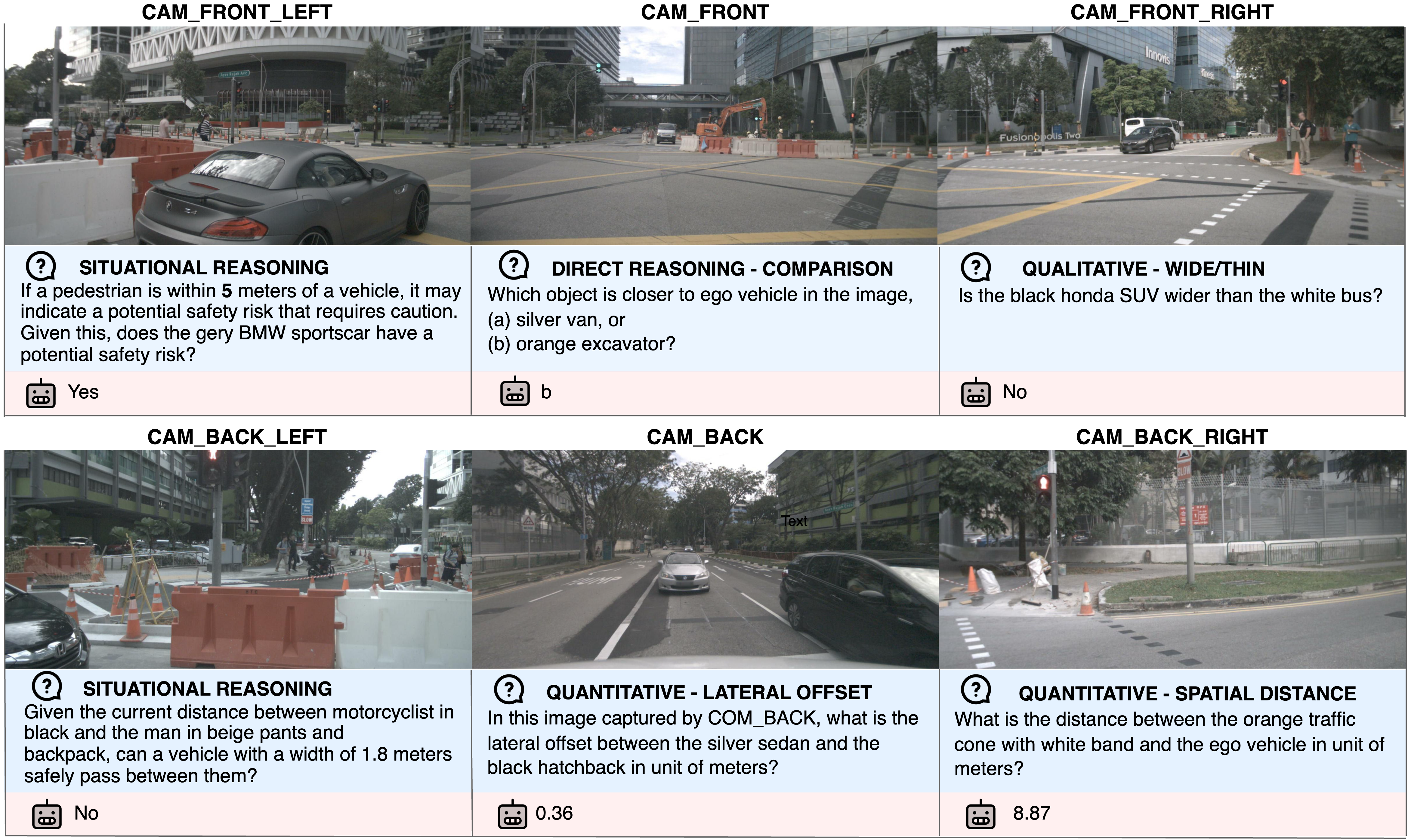}
    \caption{Example QA pairs from NuScenes-SpatialQA benchmark.}
    \label{fig:example}
\end{figure*}
The development of nuScenes-SpatialQA is centered around two key pipelines: the construction of 3D scene graphs and the generation of spatial QA pairs. The details follow below. 

\subsubsection{Raw Data}

We construct nuScenes-SpatialQA based on the nuScenes dataset~\cite{nuscenes}, a large-scale autonomous driving dataset that provides multi-modal sensor data and 3D object annotations. For this benchmark, we construct it based on the validation set of the nuScenes \textit{trainval-v1.0} split, containing 150 scenes, each with 40 key frames. Further details about the nuScenes dataset can be found in Appx.~\ref{s.1}.

\subsubsection{Auto Captioning \label{caption}}

The 3D annotations provided by NuScenes are derived from LiDAR data, providing ground truth spatial information but lacking semantic descriptions of the objects. Since VLMs rely on visual and textual input, captions are needed to establish a linguistic representation for each object, ensuring the model attends to the correct target. A well-formed caption must be clear, distinctive, and informative. To achieve this, we implement an automated captioning pipeline that systematically generates structured descriptions for each object:

\paragraph{3D-to-2D Bounding Box Projection} 

To associate each object with a corresponding image region, we project 3D bounding boxes onto the 2D image plane. Since NuScenes provides 3D object annotations rather than manually labeled 2D bounding boxes, we derive 2D bounding boxes by applying a perspective projection transformation using the calibrated camera parameters. This step ensures that each object detected in the LiDAR-based 3D space is properly localized in the camera view, facilitating accurate captioning in the subsequent stage. Further implementation details can be found in Appx.~\ref{s.2}.

\paragraph{Visibility and Resolution Filtering} 

Occluded or low-resolution objects may lack sufficient visual details for accurate captioning, leading to ambiguity. To ensure distinguishable objects for VLM recognition, we apply an initial filtering step based on visibility and bounding box size. We first remove partially or fully occluded objects using NuScenes‘ \texttt{visibility} annotations, retaining only fully visible ones. Next, we filter out objects with small 2D bounding boxes, as they may lack essential visual features. This refinement improves object selection for reliable scene representation. A global list of retained objects, including \texttt{sample\_annotation\_token}, is stored for later processing (Section~\ref{scenegraph}). Further details are provided in Appx.~\ref{s.3}.

\paragraph{Grouping Object Instances to Optimize Cropping}

Each object may appear in multiple keyframes, but only one high-quality crop is needed for caption generation. Cropping from every frame would introduce redundancy, increasing computational cost and storage. To avoid this, we group all the occurrences of each object within a scene using its \texttt{instance\_token}. This structured grouping minimizes redundancy and prepares the data for later selection. Details of grouping can be found in Appx.~\ref{s.4}.

\paragraph{Best Object View Selection} 

After grouping object appearances across frames, we now select the most informative and visually clear instance for each object to ensure high-quality caption generation. A larger bounding box generally provides a better view, but it does not always guarantee optimal clarity, as poor lighting conditions may obscure details. To refine selection, we apply a three-step process. First, we identify up to three frames where the object appears with the largest 2D bounding boxes, as larger crops tend to contain more visual details. Then, since size alone does not ensure visibility, we compute a brightness score for each candidate by averaging pixel intensity values in grayscale and select the frame with the highest score to prioritize well-lit images. Finally, we apply a 100-pixel padding around the selected crop to provide additional context for VLM captioning while avoiding interference from surrounding objects. Since padding alters brightness measurements, scores are computed before padding to ensure accurate selection.
\begin{table*}[t]
  \centering
    \setlength{\tabcolsep}{8pt}

  \begin{tabular}{>{\raggedright\arraybackslash}p{3.7cm}cc>{\centering\arraybackslash}p{1.1cm}c>{\centering\arraybackslash}p{1.1cm}cc}
    \specialrule{1.2pt}{0pt}{0pt}
        \toprule
        \multirow{2}{*}{\textbf{Benchmarks}} & \multicolumn{2}{c}{\textbf{Data Properties}} & \multicolumn{3}{c}{\textbf{Task Properties}} & \textbf{Scoring}\\
        \cmidrule(lr){2-3} \cmidrule(lr){4-7} \cmidrule(lr){7-7}
        & \makecell{Scale}&  \makecell{GT-Based\\Answer} 
        & \makecell{Spatial \\ Focus} & \makecell{Spatial \\ Evaluation Depth} 
        & \makecell{For AD} & \makecell{Model\\ Free} \\
        \midrule
        DriveLM-NuScenes~\cite{sima2024drivelm} & 0.45M & \centering ---- & \ding{55} & \centering -------- & \ding{51} & \ding{55}\\
        LingoQA~\cite{marcu2312lingoqa}  & 0.42M & \centering ---- & \ding{55} & \centering -------- & \ding{51}  & \ding{55}\\
        AutoTrust~\cite{xing2024autotrust} & 0.018M & $\triangle$ & \ding{55} & \centering -------- & \ding{51} & \ding{55}\\
        CoVLA~\cite{arai2024covla}& 6M & \ding{51} & \ding{55} & \centering -------- & \ding{51} & \ding{51}\\ 
        NuScenes-QA~\cite{qian2024nuscenesqamultimodalvisualquestion}& 0.46M & \ding{51} & $\triangle$ & Left/Right Only & \ding{51} & \ding{51}\\
        NuScenes-MQA~\cite{inoue2023nuscenesmqaintegratedevaluationcaptions}& 1.46M & \ding{51} & $\triangle$ & Distance Only & \ding{51} & \ding{51}\\
        \midrule
        VSR~\cite{liu2023visual1}& 0.01M & $\triangle$ & \ding{51} & Partial Understanding & \ding{55} & \ding{51}\\
        SpatialRGPT-Bench~\cite{cheng2024spatialrgptgroundedspatialreasoning}& 1406 & \ding{55} & \ding{51} & Understanding, Reasoning & \ding{55} & \ding{55} \\
        \midrule
        \textcolor{blue}{\ding{118}}NuScenes-SpatialQA  & 3.3M+ & \ding{52} & \ding{52} & Understanding, Reasoning & \ding{52} & \ding{52} \\
        \bottomrule
  \end{tabular}
  \caption{Comparison of NuScenes-SpatialQA with existing open-sourced autonomous driving benchmarks and spatial reasoning benchmarks. Benchmarks marked with \textcolor{blue}{\ding{118}} indicate our proposed NuScenes-SpatialQA benchmark. A \ding{51} indicates full inclusion, while \ding{55} denotes absence. The \(\triangle\) symbol represents partial inclusion.}
  \label{table.comparison}
\end{table*}

\paragraph{Generating Captions with VLM} 

With the final object crops obtained, we generate textual descriptions using LLaMA-3.2. Each cropped object image is fed into the VLM along with a structured prompt to guide caption generation. The model produces concise yet informative descriptions that capture key object attributes. Details of the VLM and prompt can be found in Appx.~\ref{s.5}.

\subsubsection{Auto Generation of 3D Scene Graphs \label{scenegraph}}

To systematically encode spatial relationships between objects, we construct a 3D scene graph that transforms raw object annotations into a structured representation. We construct an individual scene graph for each camera view in each keyframe. The construction of the scene graph enables efficient querying of spatial relations and serves as the foundation for generating spatial QA. The construction process involves defining nodes for objects and edges for their spatial relationships. 

\paragraph{Node Construction} 

Each node in the 3D scene graph represents an object and is assigned a unique \textit{node\_ID} within the graph, derived from the object's \textit{instance\_token} in NuScenes. The attributes associated with each node include its corresponding \texttt{translation} (3D coordinates $(x,y,z)$), \texttt{size} (length, width, and height), \texttt{category\_name}, and \texttt{caption}. The first three ground truth attributes are directly obtained from the NuScenes 3D annotations. The \texttt{caption} attribute is added by matching the object's \texttt{instance\_token} with the previously generated caption in \ref{caption}. This structured node representation provides a foundation for encoding spatial relationships between objects. Detailed node structure can be found in Appx.~\ref{s.6}.

\paragraph{Edge Construction} 

To encode spatial relationships between objects, we define edges in the 3D scene graph, connecting pairs of nodes within the same camera view. Specifically, each edge captures geometric relationships by computing \texttt{spatial\_distance}, \texttt{longitudinal\_offset}, \texttt{lateral\_offset}, and \texttt{relative\_bearing\_angle} between objects based on their 3D coordinates. Detailed edge structure can be found in Appx.~\ref{s.7}.

\subsubsection{Auto QA Generation}

To evaluate the spatial reasoning abilities of vision-language models, we automatically generate QA pairs based on the structured 3D scene graph. The generated questions fall into two main levels: spatial understanding and spatial reasoning.

Spatial understanding questions assess direct spatial relationships and are categorized into qualitative QA and quantitative QA. Qualitative questions evaluate relative spatial relations, such as whether one object is in front of or behind another, or whether an object is larger or smaller than another. Quantitative questions involve direct numerical estimation, requiring models to extract specific values such as distances, dimensions, or angles.

Spatial reasoning questions require higher-level inference beyond direct attribute retrieval and are categorized into direct reasoning and situational reasoning. Direct reasoning combines multiple spatial relations to derive implicit conclusions, while situational reasoning introduces contextual constraints that require the model to reason within a specific scenario.

All QA pairs are generated using predefined templates, ensuring consistency across the dataset. The QA templates for all categories can be found in Appx.~\ref{s.8}.

\subsection{Analysis \label{analysis}}

\paragraph{Statistics} 

The final benchmark consists of approximately \textbf{3.5M} total QA pairs, including approximately \textbf{2.5M} qualitative and approximately \textbf{0.6M} quantitative questions in the spatial understanding category, as well as \textbf{0.2M} reasoning-based QA covering direct and situational reasoning. These QA pairs span \textbf{6000} keyframes, each captured from 6 camera views.

\paragraph{Comparison}

To highlight the significance of NuScenes-SpatialQA, we compare it with existing open-source benchmarks in autonomous driving and spatial reasoning. As shown in Table \ref{table.comparison}, many existing benchmarks rely on depth estimation models, introducing inherent biases; in contrast, our benchmark leverages real-world ground-truth values, ensuring precise spatial alignment. Additionally, several benchmarks require external models such as GPT-4o for scoring, introducing dependencies that may obscure model performance, whereas NuScenes-SpatialQA provides a fully self-contained evaluation framework. Furthermore, while prior autonomous driving QA benchmarks include only a limited number of spatial questions, spatial reasoning benchmarks are not tailored for autonomous driving, leaving a gap in evaluating spatial reasoning within this domain. NuScenes-SpatialQA is the first large-scale ground truth-based benchmark that comprehensively evaluates spatial understanding and reasoning for autonomous driving.
\begin{table*}[ht]
  \centering
  \begin{tabular}{lcccccccc}
    \addlinespace[2pt]
    \specialrule{1.2pt}{0pt}{0pt}
    \addlinespace[2pt]
        \textbf{Models}   & \textbf{\makecell{Below/\\ Above}}
        & \textbf{\makecell{Left/\\Right}}
        & \textbf{\makecell{Front/\\Behind}}
        & \rule{0pt}{17pt} \textbf{\makecell{Large/\\Small}}
        & \textbf{\makecell{Wide/\\Thin}}
        & \textbf{\makecell{Tall/\\Short}}
        & \textbf{\makecell{Long/\\Short}}
        & \textbf{\makecell{Avg.}} \\
    \midrule
    LLaVA-v1.6~\cite{liu2023visual} & 49.78 & 49.84  & 50.14  &  59.44 &  53.84 & 55.07  &  55.73 & 53.30 \\
    Llama-3.2~\cite{grattafiori2024llama3herdmodels}& 52.12 & 51.45 & \textbf{50.84} & 60.97 & 56.78 &  53.21 & 55.49  & 54.27 \\
    blip2-flan-t5~\cite{li2023blip2bootstrappinglanguageimagepretraining}& 50.01& 50.05 & 49.87& 53.27& 50.94& 50.76& 52.00& 50.95\\
    Qwen2.5-VL~\cite{bai2025qwen2}& \textbf{53.64} & \textbf{54.55} & 50.07  &  68.58 & 60.32  & 58.62  & 61.99  & 58.02 \\
    Deepseek-vl2~\cite{wu2024deepseekvl2mixtureofexpertsvisionlanguagemodels}& 52.07 &  53.22 & 50.05  & 54.95  &  52.19 &  51.03 &  50.89 &  52.10\\ 
    \ding{71} SpatialRGPT~\cite{cheng2024spatialrgptgroundedspatialreasoning}& 53.53 & 50.91 & 50.43 & \textbf{72.25} & \textbf{64.69} & \textbf{63.60} & \textbf{65.94} & \textbf{59.79} \\ 
    \bottomrule
    \addlinespace[2pt]
    \specialrule{1.2pt}{0pt}{0pt}
    \addlinespace[2pt]
    \textbf{Models}    & \textbf{\makecell{Spatial\\Distance}}
        & \textbf{\makecell{Lateral\\Offset}}
        & \rule{0pt}{17pt} \textbf{\makecell{Longitudinal\\Offset}}
        & \textbf{\makecell{Length}}
        & \textbf{\makecell{Width}}
        & \textbf{\makecell{Height}}
        & \textbf{\makecell{Angle}}
        & \textbf{\makecell{Avg.}} \\
    \addlinespace[2pt]
    \midrule
    LLaVA-v1.6~\cite{liu2023visual} & 13.0/30.1 & 9.31/12.3 & \textbf{16.3}/\textbf{11.5} & 80.8/9.1 & \textbf{65.1}/1.4 & \textbf{84.0}/\textbf{0.4}& 11.7/168.0 \degree & \textbf{35.5}/33.3\\
    Llama-3.2~\cite{grattafiori2024llama3herdmodels}& 6.7/17.5 & 5.9/13.2 & 5.7/13.9 &72.4/3.3 & 30.4/1.7 & 5.3/46.7  &  5.3/\textbf{46.7}\degree & 16.1/\textbf{20.4} \\
    blip2-flan-t5~\cite{li2023blip2bootstrappinglanguageimagepretraining}& 6.4/17.3& 9.8/12.2& 4.3/13.8& 10.1/21.1&  8.4/2.0&  10.6/2.8& 4.7/109.2\degree& 7.8/25.5\\
    Qwen2.5-VL~\cite{bai2025qwen2}& 13.4/33.8 & 14.2/18.9  &  12.9/28.0 & \textbf{82.8}/2.2  &  14.2/2.6 & $<$0.1/45.1& \textbf{30.0}/47.4\degree  & 19.4/25.4 \\
    Deepseek-vl2~\cite{wu2024deepseekvl2mixtureofexpertsvisionlanguagemodels}& \textbf{13.7}/15.4 & 2.9/14.3  & 3.4/14.9  &  74.5/35.0 & 3.6/5.2  & 0.1/74.7  &  26.0/47.4\degree &  17.7/29.6\\ 
    \ding{71} SpatialRGPT~\cite{cheng2024spatialrgptgroundedspatialreasoning}& 7.5/\textbf{14.7}& \textbf{24.0}/\textbf{11.3}& 9.3/11.6 & 18.5/\textbf{1.4} & 11.0/\textbf{1.0} & 27.0/\textbf{0.4} & 10.7/111.7\degree & 14.6/21.7\\ 
    \bottomrule
    \addlinespace[2pt]
    \end{tabular}
    \vspace{-2mm}
    \caption{Performance on \textbf{spatial understanding} tasks in NuScenes-SpatialQA. The upper part of the table reports results on \textbf{Qualitative Spatial QA}, where values represent \textit{accuracy} (\textbf{↑}). The lower part presents results on \textbf{Quantitative Spatial QA}, where values correspond to \textit{Tolerance-based Accuracy} (\textbf{↑}) / \textit{MAE} (\textbf{↓}). Baseline marked with \ding{71} is spatial-enhanced VLM. }
    \label{tab:spatial.understanding}
\end{table*}

\begin{table*}[t]
  \centering
  \begin{tabular}{lcccc}
    \specialrule{1.2pt}{0pt}{0pt}
        \toprule
        \multirow{2}{*}{\textbf{Models}} & \multicolumn{2}{c}{\textbf{Spatial Understanding}} & \multicolumn{2}{c}{\textbf{Spatial Reasoning}} \\
        \cmidrule(lr){2-3} \cmidrule(lr){4-5}
        & \makecell{Qualitative} & \makecell{Quantitative} & \makecell{Direct Reasoning} 
        & \makecell{Situational Reasoning} \\
        \midrule
        LLaVA-v1.6-mistal-7b~\cite{liu2023visual}& 53.30 & \textbf{35.48} & 48.51 & 73.50 \\
        Llama-3.2-11B-Vision-Instruct~\cite{grattafiori2024llama3herdmodels} & 54.27 & 16.11 & 41.79 & 37.25 \\
        blip2-flan-t5-xl~\cite{li2023blip2bootstrappinglanguageimagepretraining} & 50.95 &  7.79 & 44.05  &  33.16 \\
        Qwen2.5-VL-7B-Instruct~\cite{bai2025qwen2} & 58.02 &  19.41 & \textbf{58.18} & 84.06 \\
        Deepseek-vl2-tiny~\cite{wu2024deepseekvl2mixtureofexpertsvisionlanguagemodels}& 52.10 &  17.66 &  51.76 &  \textbf{84.41} \\ 
        \ding{71} SpatialRGPT~\cite{cheng2024spatialrgptgroundedspatialreasoning}& \textbf{59.79}  &  14.59 & 45.45  & 80.77\\ 
        \bottomrule
    \end{tabular}
    \vspace{-2mm}
    \caption{Performance on \textbf{Spatial Reasoning} tasks in NuScenes-SpatialQA. The table reports \textit{Tolerance-based Accuracy} (\textbf{↑}), as defined in Section~\ref{metrics}, across different VLMs.}

  \label{tab:spatial_reasoning}
\end{table*}

%% file: sec/4_experiments.tex
\section{Experiments}

\subsection{Experimental Settings}

\paragraph{Baselines}
To evaluate the spatial understanding and reasoning capabilities of VLMs, we conduct experiments on our proposed NuScenes-SpatialQA benchmark. We select several widely used general-purpose VLMs and a state-of-the-art spatially enhanced VLM as baselines, ensuring diversity in architectures and training strategies: 
 \texttt{LLaVA-v1.6-mistral-7b}~\cite{liu2023visual}, 
 \texttt{Qwen2.5-VL-7B-Instruct}~\cite{bai2025qwen2}, 
 \texttt{blip2-flan-t5-xl}~\cite{li2023blip2bootstrappinglanguageimagepretraining}, 
 \texttt{deepseek-vl2-tiny}~\cite{wu2024deepseekvl2mixtureofexpertsvisionlanguagemodels},   
 \texttt{Llama-3.2-11B-Vision-Instruct}~\cite{grattafiori2024llama3herdmodels}, 
 \texttt{SpatialRGPT}~\cite{cheng2024spatialrgptgroundedspatialreasoning}.
These models provide a comprehensive basis for assessing spatial reasoning. Details about baselines can be found at Appx. \ref{s.9}.

\begin{table*}[t]
  \centering
  \setlength{\tabcolsep}{8pt}
  \begin{tabular}{lcccccc}
    \toprule
     \multirow{2}{*}{\textbf{Models}} & \multicolumn{2}{c}{\textbf{Spatial Understanding}} & \multicolumn{2}{c}{\textbf{Spatial Reasoning}} \\
        \cmidrule(lr){2-3} \cmidrule(lr){4-5}
        & \makecell{Qualitative} & \makecell{Quantitative} & \makecell{Direct Reasoning} 
        & \makecell{Situational Reasoning} \\
    \midrule
    LLaVA-v1.6-mistral-7b &  53.30 & \textbf{35.48} &\textbf{ 48.51} & 73.50 \\
    LLaVA-v1.6-vicuna-7b & 49.77 & 33.92 & 39.49 & 15.59  \\
    LLaVA-v1.6-vicuna-13b & 55.48 & 26.93 & 43.94 & \textbf{83.12} \\
    LLaVA-v1.6-34b & \textbf{60.79} & 20.43 & 47.57 & 80.11 \\
    \bottomrule
  \end{tabular}
  \vspace{-2mm}
  \caption{Effect of \textbf{backbone architecture} and \textbf{model scaling} on VLM performance. This table reports \textit{Tolerance-based Accuracy} (\textbf{↑}) across different model variants of LLaVA-v1.6. The first two rows compare the impact of different backbone architectures (Mistral-7B vs. Vicuna-7B). The last three rows examine the effect of model scaling.}
  \label{tab:ablation}
\end{table*}

\subsubsection{Questions and Metrics \label{metrics}}

\paragraph{Closed-Ended Questions} 

This category consists of \textit{yes-or-no} questions and \textit{multiple-choice} questions with a single correct answer. We use \textit{accuracy} as the evaluation metric, measuring the proportion of VLM responses that match the ground-truth answer.

\paragraph{Quantitative Open-Ended Questions} 

This category consists of questions that require numerical responses. While these questions allow open-ended answers, they expect a single numeric value in a predefined unit. To assess VLMs' performance on this category of questions, we use two metrics: (1) \textit{Tolerance-based Accuracy}, which measures the proportion of responses falling within the range $[75\% , 125\%]$ of the ground-truth answer; and (2) \textit{Mean Absolute Error (MAE)}, which quantifies the deviation between predictions and the ground truth.

\subsection{NuScenes-SpatialQA Benchmark Evaluation}
This section presents the evaluation of VLMs on our NuScenes-SpatialQA benchmark, focusing on two core aspects: \underline{spatial understanding} and \underline{spatial reasoning}. The following subsections detail these evaluations.

\subsubsection{Evaluating Spatial Understanding in VLMs} Spatial understanding evaluates a model’s ability to recognize and quantify spatial properties and relationships. We assess this capability through two complementary tasks: \textit{Qualitative Spatial QA} and \textit{Quantitative Spatial QA}. Table \ref{tab:spatial.understanding} reports the performance of VLMs on these tasks. 

\paragraph{Qualitative Spatial QA} The upper part of Table~\ref{tab:spatial.understanding} presents the performance of baseline VLMs on qualitative spatial understanding questions, covering seven specific categories. SpatialRGPT outperforms all baseline models, achieving the highest average accuracy and demonstrating a significant lead in size-based reasoning tasks. Qwen2.5-VL-7B-Instruct also performs competitively, excelling particularly in basic spatial relationship tasks, indicating better localization of objects in vertical and horizontal directions. However, overall accuracy remains modest, underscoring the challenges VLMs face in fine-grained spatial understanding.

\paragraph{Quantitative Spatial QA} 

The bottom part of Table~\ref{tab:spatial.understanding} reveals a trade-off between accuracy and stability in quantitative spatial reasoning. LLaVA-v1.6-mistral-7b achieves the highest accuracy, ranking first in three tasks, but also exhibits the highest MAE, indicating frequent extreme over- or under-estimations despite often falling within the correct tolerance range.
Interestingly, LLaVA-v1.6 significantly outperforms other VLMs in width and height estimation, with a particularly large gap in height. This aligns with its pretraining on GQA~\cite{liu2024improvedbaselinesvisualinstruction}, which includes height and width questions~\cite{hudson2019gqanewdatasetrealworld}, whereas Qwen-VL2.5, with its emphasis on long text, math, and coding~\cite{bai2025qwen2}, lacks spatial world knowledge.
Overall, all models struggle with quantitative spatial QA, with some tasks achieving accuracy below 0.01, underscoring the challenges VLMs face in extracting precise spatial information and generating stable outputs.

\subsubsection{Evaluating Spatial Reasoning in VLMs}

The results of the spatial reasoning capability of each baseline are shown in table \ref{tab:spatial_reasoning}. Direct Reasoning involves multi-hop reasoning based on explicit spatial relationships, while Situational Reasoning requires integrating information from multiple objects for more complex spatial inference. 

Among the baselines, Qwen2.5-VL-7B-Instruct achieves the highest accuracy in Direct Reasoning, while Deepspeck-vl2-tiny and Qwen2.5-VL-7B-Instruct excel in Situational Reasoning, reaching 84\% accuracy. Interestingly, we observed that models generally perform better on Situational Reasoning than Direct Reasoning.
This is aligned with the fact that Situational Reasoning tasks can partially leverage semantic knowledge and common spatial patterns from pretraining data, whereas Direct Reasoning requires explicit geometric understanding without relying on such priors.  
Additionally, the performance of Direct Reasoning shows a similar trend to Spatial Understanding, which may suggest that a certain level of spatial understanding serves as a foundation for spatial reasoning in VLMs.

\subsection{Ablation Study}
\subsubsection{Effect of Backbone Architecture}
The effect of architecture variation can be observed by comparing the performance of LLaVA-v1.6-mistral-7b and LLaVA-v1.6-vicuna-7b in table \ref{tab:ablation}. LLaVA-v1.6-mistral-7B is based on Mistral-7B as its foundation LLM, while LLaVA-v1.6-vicuna-7B is based on Vicuna-7B. The table demonstrates that LLaVA-v1.6-mistral-7B consistently outperforms LLaVA-v1.6-vicuna-7B across all types of QA tasks. This aligns with the fact that Vicuna-7B excels in conversational fluency~\cite{vicuna2023}, whereas Mistral-7B demonstrates better numerical and logical reasoning capabilities, enabling it to better comprehend spatial concepts and perform comparisons.

\subsubsection{Effects of Model Scaling}
The effect of scaling can be observed by comparing the performance of LLaVA-v1.6-vicuna-7b, LLaVA-v1.6-vicuna-13b, and LLaVA-v1.6-34b in table \ref{tab:ablation}. The result shows that scaling has a significant impact on qualitative understanding, with larger parameter-sized models consistently achieving higher accuracy, indicating that increased parameters enhance the model's ability to recognize spatial relationships. However, quantitative understanding does not exhibit the same trend, indicating that increasing model size alone does not necessarily enhance numerical spatial quantitative estimation capabilities. This result is aligned with the findings in~\cite{guo2024drivemllmbenchmarkspatialunderstanding}.

\begin{table}[t]
  \centering
  \begin{tabular}{lcc}
    \toprule
   \textbf{Methods} & \textbf{Vanilla} & \textbf{CoT} \\
    \midrule
    LLaVA-v1.6-mistral-7b~\cite{liu2023visual}& 61.01  &  47.27 \textcolor{red}{(-13.74)}\\
    Blip2-Flan-t5-xl~\cite{li2023blip2bootstrappinglanguageimagepretraining}& 38.60  & 39.09 \textcolor{blue}{(+0.49)} \\
    QWen2.5-VL-7B-Instruct~\cite{bai2025qwen2}& 71.12  & 63.72 \textcolor{red}{(-7.40)} \\
    DeepSeek-VL2-tiny~\cite{wu2024deepseekvl2mixtureofexpertsvisionlanguagemodels}& 68.09  & 54.82 \textcolor{red}{(-13.27)}\\
    SpatialRGPT~\cite{cheng2024spatialrgptgroundedspatialreasoning} & 63.11  & 56.85 \textcolor{red}{(-6.26)} \\
    \bottomrule
  \end{tabular}
  \vspace{-2mm}
  \caption{Effects of \textbf{CoT reasoning} on VLM performance in NuScenes-SpatialQA.}
  \label{tab:cot_transposed}
\end{table}

\subsubsection{Effect of Chain-of-Thought (CoT) Reasoning}
To investigate the impact of CoT prompting on spatial reasoning, we compare the performance of parts of VLM with and without CoT prompting. Detailed CoT Prompts can be found in Appx.~\ref{s.10}. As shown in Table~\ref{tab:cot_transposed}, surprisingly, we observe that for most models, introducing CoT prompting leads to a decline in spatial reasoning performance. This trend aligns with the findings reported in~\cite{awal2025investigatingpromptingtechniqueszero} and~\cite{yan2025multimodalinconsistencyreasoningmmir}, suggesting that explicit CoT reasoning steps may not always be beneficial for VLMs. A more effective CoT prompt needs to be designed for VLMs to enhance their reasoning capabilities.

%% file: sec/5_conclusions.tex
\section{Concluding Remarks}
In this paper, we propose \textbf{NuScenes-SpatialQA}, the first benchmark for evaluating the spatial understanding and reasoning capabilities of VLMs in autonomous driving. Using this benchmark, we assess both general-purpose and spatially enhanced VLMs. While most VLMs perform reasonably well on qualitative spatial tasks, they struggle significantly with quantitative reasoning. Spatially enhanced VLMs show improvements in qualitative understanding but no clear advantage in quantitative QA. Additionally, VLMs perform better in situational reasoning than direct geometric reasoning, indicating a reliance on world knowledge. These findings highlight persistent challenges in VLM spatial reasoning, emphasizing the need for further advancements.

\noindent \textbf{Limitations and Future Works.} While NuScenes-SpatialQA provides a systematic evaluation of spatial reasoning in VLMs, it has certain limitations. Our benchmark is constructed from the NuScenes dataset, which, while diverse, is limited to urban driving scenarios and does not cover all possible driving conditions. In future work, we aim to explore broader driving contexts and investigate methods to enhance VLM spatial reasoning performance.

\section*{Acknowledgements}

The authors would like to thank Prof. Cheng Zhang for his valuable feedback during the early stage of this work.

%% file: sec/6_supplementary.tex

\newpage

\onecolumn

\appendix

\section{Details about raw data \label{s.1}}

NuScenes \cite{nuscenes} dataset consists of 1,000 diverse urban driving scenes collected in Boston and Singapore, each lasting 20 seconds and recorded at 2 Hz. The dataset provides fully annotated 3D object detection and tracking data, featuring synchronized multi-sensor recordings from six cameras, a 32-beam LiDAR, five radars, and additional vehicle state information such as GPS and IMU.  

It provides 3D annotations for 23 object categories, including vehicles, pedestrians, traffic cones, and barriers. Each annotated object is represented by a 3D bounding box with attributes such as position, size, orientation, and visibility level. The annotations are available at 2 Hz across 1,000 urban driving scenes.

\section{Implementation Details for Auto Captioning}

\subsection{Implementation Details for 3D-to-2D Bounding Box Projection \label{s.2}}

We utilize the official nuScenes development kit (\textit{nuscenes-devkit}) for projecting 3D LiDAR bounding boxes onto the 2D image plane. Specifically, it transforms 3D bounding boxes from the LiDAR coordinate system to the camera coordinate system and applies the intrinsic camera matrix for projection. The projected points are post-processed to determine the 2D bounding box coordinates. The specific code used is from the official nuScenes repository: \url{https://github.com/nutonomy/nuscenes-devkit/blob/master/python-sdk/nuscenes/scripts/export\_2d\_annotations\_as\_json.py}

\subsection{Implementation Details for Visibility and Resolution Filter \label{s.3}}

To ensure the quality of 2D bounding boxes used in our evaluation, we apply a filtering process based on object size and visibility. Specifically, we remove bounding boxes with a \textbf{width or height $\leq 40$ pixels}, as these objects are too small to provide meaningful visual information. Additionally, we exclude objects with a \textbf{visibility token $<$ 4}, indicating that they are not fully visible in the camera view. According to the nuScenes definition, the visibility token is categorized into four levels: 1 (invisible), 2 (occluded), 3 (partially visible), and 4 (fully visible). By retaining only fully visible objects (\texttt{visibility} $= 4$), we eliminate ambiguous or heavily occluded instances, ensuring a cleaner and more reliable dataset for evaluation.

\subsection{Implementation Details for Grouping Object Instances to Optimize Cropping \label{s.4}}

In a given scene, the same object may appear across multiple keyframes. However, if we extract and store every instance of an object from each keyframe and use it as input for the VLM, this would lead to excessive memory consumption and significantly slow down the caption generation process. Additionally, since each keyframe contains multiple objects, processing every frame individually would create a large number of redundant inputs, many of which may not contribute meaningful new information. Furthermore, not all cropped images from every frame will result in high-quality captions due to variations in object visibility, occlusion, and resolution. Therefore, a more efficient strategy is needed to select representative frames for each object while maintaining high caption quality.

To efficiently select the best frames for generating high-quality captions, we group bounding boxes by object instance within each scene. Instead of processing every appearance of an object across all keyframes, we aggregate all its bounding boxes throughout the scene based on its \texttt{instance\_token}. This allows us to analyze the object's occurrences holistically and choose the most representative frames for caption generation.

\subsection{Implementation Details for Generating Captions with VLM \label{s.5}}

We select \textbf{Llama-3.2-11b-Instruct} as the VLM for caption generation. 
For each selected object instance, we extract its cropped image and feed it into the VLM with the following prompt:

\begin{tcolorbox}[
    colback=gray!5!white, colframe=gray!75!black, 
    title=Prompt for Caption Generation,
    fonttitle=\bfseries
]
\small
\ttfamily
\raggedright
Provide a short noun phrase captioning \{category\_name\} in the center of the image,  
such as 'black sedan with red logo' or 'man in a blue t-shirt and jeans'.  
The response must be a phrase only. Do NOT include full sentences or extra descriptions.
\end{tcolorbox}

This prompt encourages the model to generate concise yet discriminative captions.

\section{Implementation Details for Auto Generation of 3D Scene Graphs }

\subsection{Node Structure \label{s.6}}

The node structure in our scene graph is represented using the following JSON format:

\begin{tcolorbox}[
    colback=gray!5!white, colframe=gray!75!black, 
    title=Node Structure: JSON Representation,
    fonttitle=\bfseries,
    boxsep=1mm,
    left=1mm,
    right=1mm 
]
\begin{lstlisting}[basicstyle=\ttfamily\footnotesize, breaklines=true]
{
    "node_id": "a721d524937f4a228fa6aac3296fb3bc",
    "attributes": {
        "category_name": "human.pedestrian.adult",
        "translation": {
            "x": 286.706,
            "y": 926.831,
            "z": 1.176
        },
        "size": {
            "length": 1.095,
            "width": 0.695,
            "height": 1.78
        },
        "caption": "the man in tan t-shirt and jeans"
    }
}
\end{lstlisting}
\end{tcolorbox}

Here, \texttt{node\_id} corresponds to the instance token, which uniquely identifies an object across different frames. The \texttt{attributes} field contains essential properties of the object, including its \texttt{category\_name}, \texttt{translation} (3D coordinate), \texttt{size} (physical dimension), and a \texttt{caption} generated by the auto-captioning process.

\subsection{Edge Structure \label{s.7}}

The edge structure in our scene graph is represented using the following JSON format:

\begin{tcolorbox}[
    colback=gray!5!white, colframe=gray!75!black, 
    title=Edge Structure: JSON Representation,
    fonttitle=\bfseries,
    boxsep=1mm, 
    left=1mm, 
    right=1mm 
]
\begin{lstlisting}[basicstyle=\ttfamily\footnotesize, breaklines=true]
{
    "edge_id": "296fb3bc_b43a1a15",
    "from": "a721d524937f4a228fa6aac3296fb3bc",
    "to": "069a7d8902d14560b1890064b43a1a15",
    "spatial_distance": 1.25,
    "longitudinal_offset": 0.65,
    "lateral_offset": 1.06,
    "relative_bearing_angle": -121.66
}
\end{lstlisting}
\end{tcolorbox}

Each edge in the graph encodes spatial relationships between objects. The attributes are defined as follows: \texttt{edge\_id} is a unique identifier for the edge, while \texttt{from} and \texttt{to} represent the unique IDs of the connected nodes (objects). \texttt{spatial\_distance} denotes the Euclidean distance between the two objects in meters. \texttt{longitudinal\_offset} refers to the displacement along the ego vehicle’s heading direction, whereas \texttt{lateral\_offset} indicates the perpendicular displacement relative to the ego vehicle’s heading. Finally, \texttt{relative\_bearing\_angle} represents the angle (in degrees) from the \texttt{from} node to the \texttt{to} node in the ego vehicle’s reference frame.

\section{Details for QA Template \label{s.8}}







\begin{tcolorbox}[
    colback=gray!5!white, colframe=gray!75!black, 
    title=QA Template: Qualitative,
    fonttitle=\bfseries,
    boxsep=1mm, 
    left=1mm, 
    right=1mm 
]

\begin{itemize}
    \item \textbf{Q:} Is \textit{\{\textit{object\_1}\}} above \textit{\{\textit{object\_2}\}}? \\
          \textbf{A:} yes / no
          \vspace{3mm}
    \item \textbf{Q:} Is \textit{\{\textit{object\_1}\}} below \textit{\{\textit{object\_2}\}}? \\
          \textbf{A:} yes / no
          \vspace{3mm}
    \item \textbf{Q:} Is \textit{\{\textit{object\_1}\}} to the left of \textit{\{\textit{object\_2}\}}? \\
          \textbf{A:} yes / no
          \vspace{3mm}
    \item \textbf{Q:} Is \textit{\{\textit{object\_1}\}} to the right of \textit{\{\textit{object\_2}\}}? \\
          \textbf{A:} yes / no
          \vspace{3mm}
    \item \textbf{Q:} Is \textit{\{\textit{object\_1}\}} in front of \textit{\{\textit{object\_2}\}}? \\
          \textbf{A:} yes / no
          \vspace{3mm}
    \item \textbf{Q:} Is \textit{\{\textit{object\_1}\}} behind \textit{\{\textit{object\_2}\}}? \\
          \textbf{A:} yes / no
          \vspace{3mm}
    \item \textbf{Q:} Is \textit{\{\textit{object\_1}\}} larger than \textit{\{\textit{object\_2}\}}? \\
          \textbf{A:} yes / no
          \vspace{3mm}
    \item \textbf{Q:} Is \textit{\{\textit{object\_1}\}} smaller than \textit{\{\textit{object\_2}\}}? \\
          \textbf{A:} yes / no
          \vspace{3mm}
    \item \textbf{Q:} Is \textit{\{\textit{object\_1}\}} longer than \textit{\{\textit{object\_2}\}} in length? \\
          \textbf{A:} yes / no
          \vspace{3mm}
    \item \textbf{Q:} Is \textit{\{\textit{object\_1}\}} shorter than \textit{\{\textit{object\_2}\}} in length? \\
          \textbf{A:} yes / no
          \vspace{3mm}
    \item \textbf{Q:} Is \textit{\{\textit{object\_1}\}} taller than \textit{\{\textit{object\_2}\}} in height? \\
          \textbf{A:} yes / no
          \vspace{3mm}
    \item \textbf{Q:} Is \textit{\{\textit{object\_1}\}} shorter than \textit{\{\textit{object\_2}\}} in height? \\
          \textbf{A:} yes / no
          \vspace{3mm}
    \item \textbf{Q:} Is \textit{\{\textit{object\_1}\}} wider than \textit{\{\textit{object\_2}\}}? \\
          \textbf{A:} yes / no
          \vspace{3mm}
    \item \textbf{Q:} Is \textit{\{\textit{object\_1}\}} thinner than \textit{\{\textit{object\_2}\}}? \\
          \textbf{A:} yes / no
\end{itemize}

\end{tcolorbox}

\begin{tcolorbox}[
    colback=gray!5!white, colframe=gray!75!black, 
    title=QA Template: Quantitative,
    fonttitle=\bfseries,
    boxsep=1mm, 
    left=1mm, 
    right=1mm, 
    breakable
]

\begin{itemize}
    \item \textbf{Q:} In this image captured by \textit{\{camera\}}, what is the distance between \textit{\{object\_1\}} and \textit{\{object\_2\}}? \\
          \textbf{A:} (numeric value)
          \vspace{3mm}
    \item \textbf{Q:} In this image captured by \textit{\{camera\}}, what is the distance between the ego vehicle and \textit{\{object\}}? \\
          \textbf{A:} (numeric value)
          \vspace{3mm}
    \item \textbf{Q:} In this image captured by \textit{\{camera\}}, what is the longitudinal offset between the ego vehicle and \textit{\{object\}}? \\
          \textbf{A:} (numeric value)
          \vspace{3mm}
    \item \textbf{Q:} In this image captured by \textit{\{camera\}}, what is the longitudinal offset between \textit{\{object\_1\}} and \textit{\{object\_2\}}? \\
          \textbf{A:} (numeric value)
          \vspace{3mm}
    \item \textbf{Q:} In this image captured by \textit{\{camera\}}, what is the lateral offset between the ego vehicle and \textit{\{object\}}? \\
          \textbf{A:} (numeric value)
          \vspace{3mm}
    \item \textbf{Q:} In this image captured by \textit{\{camera\}}, what is the lateral offset between \textit{\{object\_1\}} and \textit{\{object\_2\}}? \\
          \textbf{A:} (numeric value)
          \vspace{3mm}
    \item \textbf{Q:} What is the length of \textit{\{object\}} in the image? \\
          \textbf{A:} (numeric value)
          \vspace{3mm}
    \item \textbf{Q:} What is the width of \textit{\{object\}} in the image? \\
          \textbf{A:} (numeric value)
          \vspace{3mm}
    \item \textbf{Q:} What is the height of \textit{\{object\}} in the image? \\
          \textbf{A:} (numeric value)
          \vspace{3mm}
    \item \textbf{Q:} What is the relative bearing angle of \textit{\{object\_2\}} with respect to \textit{\{object\_1\}}? \\
          \textbf{A:} (numeric value)
\end{itemize}
\end{tcolorbox}

\begin{tcolorbox}[
    colback=gray!5!white, colframe=gray!75!black, 
    title=QA Template: Direct Reasoning,
    fonttitle=\bfseries,
    boxsep=1mm, 
    left=1mm, 
    right=1mm, 
    breakable
]

\begin{itemize}
    \item \textbf{Question:} From the given options, which object is the closest to \textit{\{object\}}? \\
          (a) \textit{\{object\_1\}} \\
          (b) \textit{\{object\_2\}} \\
          (c) \textit{\{object\_3\}} \\
          (d) \textit{\{object\_4\}} \\
          \textbf{Answer:} a / b / c / d
          \vspace{3mm}
          
    \item \textbf{Question:} This image is captured by one of the onboard cameras mounted on a vehicle. From the given options, which object is the closest to the ego vehicle? \\
          (a) \textit{\{object\_1\}} \\
          (b) \textit{\{object\_2\}} \\
          (c) \textit{\{object\_3\}} \\
          (d) \textit{\{object\_4\}} \\
          \textbf{Answer:} a / b / c / d
          \vspace{3mm}

    \item \textbf{Question:} From the given options, which object is the largest in terms of overall size? \\
          (a) \textit{\{object\_1\}} \\
          (b) \textit{\{object\_2\}} \\
          (c) \textit{\{object\_3\}} \\
          (d) \textit{\{object\_4\}} \\
          \textbf{Answer:} a / b / c / d
          \vspace{3mm}

    \item \textbf{Question:} Which object is closer to \textit{\{object\_c\}} in the image? \\
          (a) \textit{\{object\_1\}} \\
          (b) \textit{\{object\_2\}} \\
          \textbf{Answer:} a / b
          \vspace{3mm}

    \item \textbf{Question:} This image is captured by one of the onboard cameras mounted on a vehicle. Which object is closer to the ego vehicle? \\
          (a) \textit{\{object\_1\}} \\
          (b) \textit{\{object\_2\}} \\
          \textbf{Answer:} a / b
          \vspace{3mm}

    \item \textbf{Question:} Are there any pedestrians or vehicles within 5 meters of \textit{\{object\}} in the image? \\
          \textbf{Answer:} yes / no
          \vspace{3mm}

    \item \textbf{Question:} This image is captured by one of the onboard cameras mounted on a vehicle. Are there any pedestrians or vehicles within 10 meters of the ego vehicle? \\
          \textbf{Answer:} yes / no
          \vspace{3mm}

    \item \textbf{Question:} Are there any vehicles in the image with a width greater than 2 meters? \\
          \textbf{Answer:} yes / no

\end{itemize}

\end{tcolorbox}

\begin{tcolorbox}[
    colback=gray!5!white, colframe=gray!75!black, 
    title=QA Template: Situational Reasoning,
    fonttitle=\bfseries,
    boxsep=1mm, 
    left=1mm, 
    right=1mm, 
    breakable
]

\begin{itemize}
    \item \textbf{Question:} This image is captured by one of the onboard cameras mounted on a vehicle. 
          In autonomous driving, it is crucial to detect potential safety risks, 
          especially when pedestrians are too close to vehicles. If a pedestrian is within 10 meters 
          of a vehicle, it may indicate a potential hazard that requires caution. Given this, 
          does the ego vehicle have a potential safety risk due to nearby pedestrians? \\
          \textbf{Answer:} yes / no
          \vspace{3mm}

    \item \textbf{Question:} Assume the distance between \textit{\{object\_1\}} and \textit{\{object\_2\}} 
          is decreasing at 2 meters per second. Will they collide within 5 seconds? \\
          \textbf{Answer:} yes / no
          \vspace{3mm}

    \item \textbf{Question:} This image is captured by one of the onboard cameras mounted on a vehicle. 
          Assume the ego vehicle is moving forward while all other objects remain stationary. 
          Will there be a moment when \textit{\{object\_1\}} occludes \textit{\{object\_2\}}, causing 
          \textit{\{object\_2\}} to become invisible from the ego vehicle's perspective? \\
          \textbf{Answer:} yes / no
          \vspace{3mm}

    \item \textbf{Question:} Assume there is a bridge ahead with a maximum clearance height of 2 meters. 
          Any vehicle taller than this cannot safely pass under. Given this assumption, 
          is there any vehicle in the current scene unable to pass under the bridge? \\
          \textbf{Answer:} yes / no
          \vspace{3mm}

    \item \textbf{Question:} Assume there is a parking spot measuring \textit{\{spot\_length\}} meters in length and 
          \textit{\{spot\_width\}} meters in width. Considering that a vehicle needs at least 
          \textit{\{clearance\}} meters of clearance on both the front/back and left/right sides, 
          can \textit{\{object\}} in the image fit into this parking spot? \\
          \textbf{Answer:} yes / no
          \vspace{3mm}

    \item \textbf{Question:} Given the current distance between \textit{\{object\_1\}} and \textit{\{object\_2\}}, 
          can a vehicle with a width of \textit{\{vehicle\_width\}} meters safely pass between them? \\
          \textbf{Answer:} yes / no

\end{itemize}

\end{tcolorbox}

\section{Details for Baselines \label{s.9}}

\begin{table*}[t]
  \centering
  \begin{tabular}{lcccccc}
    \toprule
    \textbf{Models}   & \makecell{LLaVA-v1.6}
        & \makecell{Llama-3.2}
        & \makecell{blip2} 
        & \rule{0pt}{17pt} \makecell{Qwen2.5-VL} 
        & \makecell{Deepseek-vl2}
        & \makecell{SpatialRGPT} \\
    \midrule
    Backbone & Mistral-7B & \makecell{Llama 3.2} & Flan-T5-XL & Qwen2.5 & \makecell{MoE\\Transformer} &  \makecell{pre-trained \\ OpenAI  CLIP-L} \\
    Parameter Size & 7B & 11B & 3B & 7B & 3B &  8B \\
    \bottomrule
  \end{tabular}
  \caption{Baseline}
  \label{tab:baselinesappx}
\end{table*}

Please refer to Table \ref{tab:baselinesappx} for the backbone and parameter size of baseline VLMs.

\section{Details for CoT \label{s.10}}

\begin{tcolorbox}[
    colback=gray!5!white, colframe=gray!75!black, 
    title=Prompt for CoT Reasoning,
    fonttitle=\bfseries
]
\small
\ttfamily
\raggedright
"You are given a question about spatial relationships in an autonomous driving scene. Think step by step before answering. First, analyze the spatial arrangement of objects based on the given context. Then, determine the correct answer based on your reasoning. Finally, provide your answer in the following format:\\
Reasoning: (Step-by-step explanation)\\
Answer: (Yes/No) / (A/ B) / (A/ B/ C/ D) \textit{(according to question type)} \\
Question: \textit{\{question\}}"
\end{tcolorbox}

\section{Broader Impact and Ethics Statement}
\subsection{Broader Impact Statement}
NuScenes-SpatialQA provides a benchmark to evaluate the spatial reasoning capabilities of VLMs. Accurate spatial understanding is crucial for AI applications in autonomous driving, robotic navigation, and general visual perception. By systematically assessing VLMs' ability to interpret spatial relationships, our work helps identify limitations and guide improvements in AI-driven spatial reasoning.

\subsection{Ethics Statement}
Our research emphasizes fairness, transparency, and reliability. The benchmark is built on publicly available data while ensuring privacy and unbiased evaluation. We acknowledge the challenges of spatial reasoning in AI and advocate for responsible model development to minimize errors and unintended biases in real-world applications.